%% file: ms.tex
\documentclass[sigconf]{acmart}

\usepackage{booktabs} 
\usepackage{graphicx}
\usepackage{todonotes}
\usepackage{multicol}
\usepackage[ruled]{algorithm2e}
\usepackage{bm}  
\usepackage{arydshln}
\usepackage{tikz}
\usepackage{tikzsymbols}
\usepackage{hyperref}
\usepackage{natbib}
\usepackage{soul}
\let\checkmark\undefined
\usepackage{dingbat}
\usepackage{amsthm}
\usepackage{bm}  
\usepackage{accents}
\usepackage{enumerate}
\usepackage{lscape}

\usepackage[english]{babel}
\usepackage{blindtext}
\usepackage{etoolbox}
\makeatletter
\patchcmd{\maketitle}{\@copyrightspace}{}{}{}
\makeatother
\renewcommand\footnotetextcopyrightpermission[1]{} 
\usetikzlibrary{shapes, arrows, fit, positioning, bayesnet, snakes, decorations, decorations.markings,decorations.pathmorphing}
\tikzstyle{vecArrow} = [thick, decoration={markings,mark=at position 1 with {\arrow[semithick]{open triangle 60}}}, double distance=1.4pt, shorten >= 5.5pt, preaction = {decorate},postaction = {draw,line width=1.4pt, white,shorten >= 4.5pt}]
\tikzstyle{innerWhite} = [semithick, white,line width=1.4pt, shorten >= 4.5pt]

\newcommand{\occ}{{\em OCC}\xspace}

\begin{document}

\title{Improving Humanness of Virtual Agents and Users' Cooperation through Emotions}  

\author{Moojan Ghafurian}
\affiliation{%
  \institution{School of Computer Science
  University of Waterloo\\
  Waterloo, Ontario, N2L 3G1}
}
\email{moojan@uwaterloo.ca}

\author{Neil Budnarain}
\affiliation{%
  \institution{School of Computer Science
  University of Waterloo\\
  Waterloo, Ontario, N2L 3G1}
}
\email{bbudnarain@edu.uwaterloo.ca}

\author{Jesse Hoey}
\affiliation{%
  \institution{School of Computer Science
  University of Waterloo\\
  Waterloo, Ontario, N2L 3G1}
}
\email{jhoey@cs.uwaterloo.ca}

\begin{abstract} 
In this paper, we analyze the performance of an agent developed according to a well-accepted appraisal theory of human emotion with respect to how it modulates play in the context of a social dilemma. We ask if the agent will be capable of generating interactions that are considered to be more human than machine-like. We conduct an experiment with $117$ participants and show how participants rate our agent on dimensions of human-uniqueness (which separates humans from animals) and human-nature (which separates humans from machines). We show that our appraisal theoretic agent is perceived to be more human-like than baseline models, by significantly improving both human-nature and human-uniqueness aspects of the intelligent agent. We also show that perception of humanness positively affects enjoyment and cooperation in the social dilemma. 

\end{abstract}


\maketitle


\input{sections/IntoandBackground.tex}

\input{sections/agent.tex}
\input{sections/OCC.tex}

\input{sections/experiment.tex}

\input{sections/discussions.tex}

\input{sections/conclusionFuture.tex}

\input{sections/acknowledgements.tex}


\bibliographystyle{ACM-Reference-Format} 
\bibliography{bibliography}  

\end{document}

%% file: sections/IntoandBackground.tex
\section{Introduction}

In this paper, we investigate a theory of emotion as the method for generating artificially intelligent agents that seem more human-like. We argue that much Artificial Intelligence  (AI) research has focused on building intelligence based on individual attributes that non-human animals do not possess, but that machines inherently do possess.

However, human intelligence also requires emotional attributes and social support, attributes that machines do not possess, whereas non-human animals do.

The concept of ``human-ness'' has seen much debate in social psychology, particularly in relation to work on stereotypes and de-humanization or infra-humanisation~\cite{Demoulin2004}. Since Mori wrote about the ``uncanny valley''~\cite{Mori1970}, artificial intelligence researchers (particularly in the fields of affective computing and human-computer/robot interaction) have shown an interest in this issue. In a comprehensive series of experiments, Haslam {\em et al.}~\cite{Haslam2008} examined how people judge others as human or non-human (dehumanized).  In their model, \textit{humanness} is broken down into two factors. First, \textit{Human Uniqueness} (HU) distinguishes humans from animals (but not necessarily from machines). Second, \textit{Human Nature} (HN) distinguishes humans from machines (but not necessarily from animals). Human uniqueness traits are civility, refinement, moral sensibility, rationality and maturity, as opposed to lack of culture, coarseness, amorality, irrationality and childlikeness. Human nature traits are emotionality, warmth, openness, agency (individuality), and depth, as opposed to inertness, coldness, rigidity, passivity and superficiality. Thus, while one can imagine both humans and machines having \textit{human uniqueness} traits, animals would not tend to have these (they are coarse, amoral, etc). Similarly, while one can imagine both humans and animals having \textit{human nature} traits, machines would not tend to have these (they are inert, cold, rigid, etc). While much research in AI is trying to build machines with HU traits (thus separating AI from animals), there is much less work on trying to build machines with HN traits (thus separating AI from machines). While both problems present challenges, the former problem is already ``solved'' to a certain degree by simply having a machine in the first place, as humans and machines are on the same side of the human uniqueness divide anyway. The latter problem is more challenging as the human nature dimension is exactly the dimension on which machines differ most from humans.

Several studies have verified that humanness and emotions of virtual agents can affect people's behaviour and strategies~\cite{chowanda2016playing,de2010influence}. For example, Chowanda {\em et al.}~\cite{chowanda2016playing} captured players' emotions through their facial expressions and showed that Non-Player Characters that have personalities and are capable of perceiving emotions can enhance players' experience in the game. Further, Nonverbal behaviour such as body gesture and gaze direction affects perception of cooperativeness of an agent~\cite{strassmann2016effect}. 

In this paper, we use an appraisal-based emotional model in the same spirit as EMA~\cite{Gratch2004}, where emotional displays are made using the Ortony, Clore and Collins (OCC) model~\cite{Ortony1988}, and a set of coping rules are implemented to map the game history augmented with emotional appraisals to actions for the virtual agent. We refer to the agent based on this model as the \occ agent.

The \occ agent uses emotions to generate expectations about future actions~\cite{Fridja10,Zajonc2000}. That is, it sees emotions as being related to an agent's assessments of what is going to happen next, both within and without the agent's control. The \occ agent computes expectations with respect to the {\em denotative} meaning (or {\em causal interpretation}~\cite{Gratch2004}) of the situation, and these expectations are mapped to emotion labels. The generated emotions are then used with a set of {\em coping} rules to adjust future actions.

We present results from a study involving $N=117$ participants who played a simple social dilemma game with a virtual agent named ``Aria''. The game was a variation of Prisoner's Dilemma (PD), in which each player could either give two coins or take one coin from a common pile. Players could maximize their returns by defecting while their partner cooperated, and although the Nash equilibrium is mutual defection, the players can jointly maximize their scores through mutual cooperation. The participants were awarded a bonus according only to their total score in the game, and so had incentive to cooperate as much as possible. Participants played a series of $25$ games in a row, and then answered a questionnaire on how they felt about Aria on dimensions of human-uniqueness and human-nature taken from~\cite{Haslam2008}.  The participants were evenly split into three conditions that differed only in the emotional displays. The virtual human, Aria, displayed facial expressions and uttered canned sentences that were consistent with the game context and the emotional state given the condition. One condition was based on the \occ model, while two were baselines, one with randomly selected emotions, and one with no emotions. Our hypothesis was that the \occ agents would show more human-nature traits than the baselines.

The primary contribution of this paper is to evaluate how appraised emotions relate to two important dimensions of humanness, and to investigate the impact of appropriate emotion modeling on perceived humanness of a virtual agent and users' cooperation. Secondary contributions are a complete description of the prisoner's dilemma in terms of OCC emotions and a demonstration in a simple environment.

\section{Related Work}
Affective computing (AC) has formed as a sub-discipline of artificial intelligence seeking to understand how human emotions can be computationally modeled and implemented in virtual agents. While much current work is focused on social signal processing, the emphasis is on the detection, modeling and generation of signals relating to social interactions without considering the control mechanisms underlying the function of emotion~\cite{vinciarelli2012bridging}. In a broad survey~\cite{reisenzein2013computational}, Reisenzein {\em et al.} define emotional functions as being {\em informational}, {\em attentional}, and {\em motivational}, but point to a lack of explicit mechanisms for computational implementation.

Much of the work in AC on the function of emotions has focused on appraisal theories of emotion, as these give clear rules mapping denotative states to emotions and show a clear path for implementation. The appraisal model of Ortony, Clore and Collins (OCC)~\cite{Ortony1988} describes appraisals of the consequences of events and the actions of agents in relation to self or other, and is perhaps the most well used in AC. Scherer's component process model~\cite{Scherer2001} breaks emotion processes down into five components of appraisal, activation, expression, motivation, and feeling. 

Emotion is proposed as a facilitator of learning and as a mechanism to signal, predict, and select forthcoming action, although no precise definitions of such a mechanism is proposed. Ortony, Norman and Revelle~\cite{Ortony2005} describe a general cognitive architecture that incorporates affect at three levels of processing (reactive, routine and reflective) and four domains of processing (affect, motivation, cognition, and behaviour).  The reactive level is relegated to a lower-level (e.g. hardware on a robot) process that encodes motor programs and sensing mechanisms. They claim that appraisal and personality arise primarily at the routine level, but this is somewhat contradictory since appraisals involve reasoning about long-term goals (for example), something that is defined as being only at the reflective level in~\cite{Ortony2005}.

Efforts at integrating appraisal models in artificial agents started with Elliott's use of an OCC model augmented with ``Love'', ``Hate'' and ``Jealousy'' to make predictions about human's emotional ratings of semantically ambiguous storylines and to drive a virtual character~\cite{Elliott1998}. This was followed by the work of Gratch~\cite{Gratch2000,Gratch2004}, and OCC models were integrated with probabilistic models for intelligent tutoring applications in~\cite{Sabourin2011,Conati2009}.  A general-purpose game engine for adding emotions is described in~\cite{Gamygdala2014}.

The role of affect in decision making has long been problematic, and focus on bounded rationality following Simon's articulation of emotions as interrupts to cognitive processing~\cite{Simon67} has been pursued by many authors~\cite{Mellers99,Lisetti2002,Antos2011}. Usually, these approaches take the stance that the agent is still acting on rational and decision theoretic principles, but has a ``modified'' utility function~\cite{BenabouTirole2006}, with some tuning parameter that trades off social normative effects modeled as intrinsic rewards with the usual extrinsic rewards. The {\em affect-as-cognitive-feedback} approach~\cite{Huntsinger2014} slightly rearranges things, and proposes that affect serves as a reward signal for the default cognitive processing mechanisms associated with the situation. The BayesACT model uses the concepts of identity to frame action based on affect alone, considering it to be a key component of the social glue that enables collaboration~\cite{Hoey2016}.

Expected and immediate emotions have been related to expected utility and modify action choices accordingly~\cite{Lerner2003,ghafurian2018impatience}. Other approaches have attempted to reverse-engineer emotion through Reinforcement Learning (RL) models that interpret the antecedents of emotion as aspects of the learning and decision-making process, but relegate the function of emotion to characteristics of the RL problem~\cite{Moerland2017}. Many of these approaches borrow from behavioral economics and cognitive science to characterize the consequences of emotion in decision making and integrate this knowledge as ``coping rules''~\cite{Gratch2004}, ``affect heuristics''~\cite{Slovic2007}, or short-circuit ``impulsive behaviours''~\cite{Insua2017} to characterize or influence agents' behaviour.

Investigations into the role of emotions in modifying behaviours in a PD have looked at how {\em disappointment} and {\em anger} can be used to promote forgiveness and retaliation, respectively~\cite{Wubben2009}. Guilt has also been examined and human estimates of behaviour conditioned on facial expressions or written descriptions of this emotion have been shown to align with appraisal theoretic predictions~\cite{deMelo2012}. Human and bot play has been analyzed using BayesACT~\cite{JungHoey2016}, who showed that BayesACT could replicate some aspects of human play in PD. Poncela {\em et al.}~\cite{Poncela2016} demonstrated how humans split into different player types when playing PD, which are loosely defined on two axes of optimism-pessimism and envy-trust.

Finally, general personalities based on emotional factors of tolerance and responsiveness, and the emotion of admiration were considered in PD-like situations~\cite{Lloydkelly2012}. 

Work on negotiations is based in similar theoretical ideas~\cite{VanKleef2004}, where expressions of anger (or happiness) signal that an agent's goals are higher (or lower) than what the agent currently perceives as the expected outcome.

These findings about human-human negotiations are replicated in a human-agent study, in which verbal and non-verbal displays of anger and happiness are also compared~\cite{deMelo2011}.

Finally, the same findings were partially replicated in real (unscripted) human-human negotiations~\cite{Stratou2015}, where it was reported that negative displays increased individual gain, but led to worse longer-term outcomes.

%% file: sections/agent.tex
\begin{figure}[t]
  \begin{center}
    \includegraphics[width=0.75\linewidth]{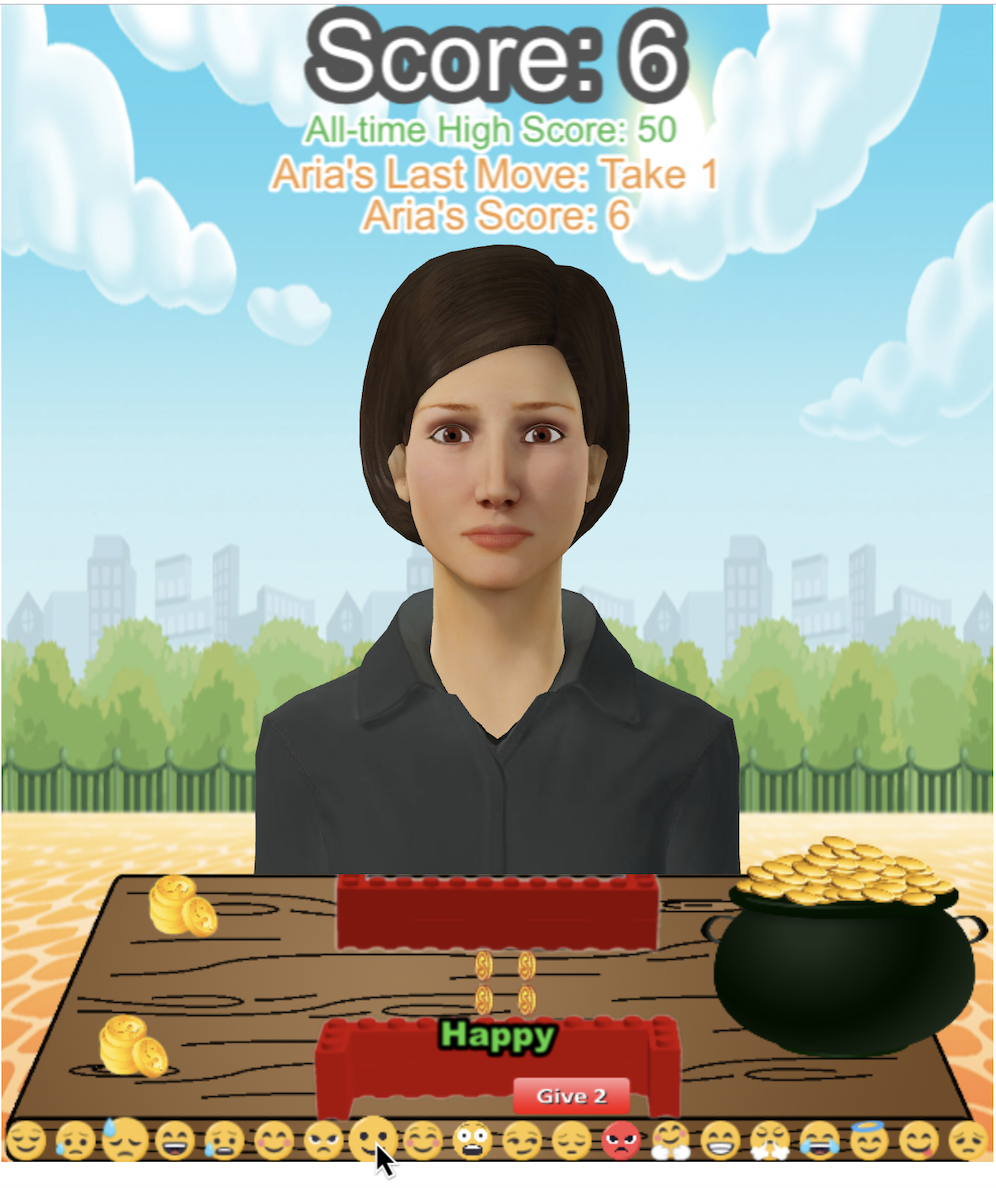}
    \end{center}
\caption{\label{fig:aria} An example of the game setting. The two coin piles on the left show the number of coins that the participant and Aria have earned so far in the game. Currently, the player has chosen to give 2 and look "happy" for the next round. Aria's current emotion is "sorry".}
  \end{figure}

\section{Game Play}
Our prisoner's dilemma game setting was implemented using Phaser (open source HTML5 game framework) and is shown in Figure~\ref{fig:aria}. The female Virtual Human was originally developed for speech and language therapy~\cite{VanVuuren2014} and has been also used in assistive technology applications~\cite{Malhotra2016}. Here, we call her Aria. 
In the game, Aria, the virtual opponent, is sitting in front of the participants. On the right, there is a large pile of coins that the players use. On the left there are two piles of coins, one pile representing the coins that Aria has received so far and the other showing participants' coins. In each round, the players' decisions are hidden from each other (hidden behind the red barriers). They can both choose between giving the other player two coins, or taking one for themselves. After the participants choose their actions, they will see the emoji list (and an emotion word describing each emoji) at the bottom of the page. After deciding on the action, participants are asked to choose one emoji based on their emotion. Upon selecting both action and emotion, the results will be revealed (coins will appear on the table and will move to the players' piles) and both players will see the other player's emotion. Aria's emotions are reflected through her facial expressions and utterances, and players' emotions are shown via the emoji that they have selected.

\subsection{Emotion Dictionary}
We use a set of 20 emotions compiled from the OCC model (see Section~\ref{sec:occ}) and these are mapped to a three dimensional emotion space with dimensions of Evaluation (E), Potency (P) and Activity (A) which is sometimes known as the VAD model where the terms are Valence (E), Activity (A) and Dominance (P)~\cite{Osgood1975}. 

We use an emotion dictionary consisting of a set of 300 emotion words rated on E,P and A by 1027 undergraduates at the University of Indiana in 2002-2004~\cite{IndianaData2006}. We manually find the same or a synonymous word in the dictionary for each of the 20 OCC emotion words. We also select a set of 20 emojis, one per OCC emotion word, to be used in the game play as described above. 

An important emotion in PD is regret. If a player defects, but regrets, it is quite different than when a player defects but shows no regret. In the following, we define regret as any of the four OCC emotions "remorse", "distress", "shame" or "fears-confirmed". Only this last term does not have a direct equivalent in the Indiana dictionary, and for it we find the term "heavy hearted" (E -1.03; P: -0.55; A: -1.15).

\subsection{Facial Expressions}
 Aria's facial expressions are generated with three controls that map to specific sets of facial muscles.  We refer to this three dimensional space of control as the ``HSF''space: (1) Happy/Sad, (2) Surprise/Anger, and (3) Fear/Disgust. 
 A setting of these three controls yields a specific facial expression by virtually moving the action units in the face corresponding to that emotion by an amount proportional to the control. For example, "happy" is expressed with AU6 + AU12: cheek raiser and lip corner puller. Although the virtual human's face can be controlled by moving individual muscles, like the inner eyebrow raise, groups of these are highly correlated and move in recognizable patterns. Therefore, these three dimensions of musculature movement are deemed sufficient.

To map from an emotion label from our set of 20, first the emotion word is mapped to EPA space using the Indiana dictionary, and then the distance to each end-point of the HSF space is computed using the EPA ratings shown below, also from the Indiana dictionary, and these distances are used to set the HSF controls directly.

\begin{small}
\begin{center}
 \begin{tabular}{lcl}
 Emotion & Symbol & E,P,A \\ \hline
 happy / sad & $h_+$ / $h_-$ & (3.45,2.91,0.24) / (-2.38,-1.34,-1.88)\\
 surprise / anger & $h_+$ / $h_-$& (1.48,1.32,2.31) / ( -2.03,1.07,1.80)\\ 
fear / disgust & $h_+$ / $h_-$ & (-2.41,-0.76,-0.68) / (-2.57,0.27,0.43)\\
\hline
 \end{tabular}
 \end{center}
\end{small}

 Aria has a normal "quiescent" state in which she blinks and slightly moves her head from side to side in a somewhat random way. The emotions are applied for 10.5 seconds and the face smoothly transitions to a weaker representation of the same emotion and the quiescent state. 

\subsection{Utterances}
\label{sec:utterance}
Aria also utters sentences from 8 predefined sets, one for each combination of agent action, human action, and binary indicator of valence (E) being positive. Speaking and facial expressions are possible at the same time. Lip movement during speech is based on a proprietary algorithm.

An embedding (vector) for each emotion label is computed using the pretrained Word2Vec model which was trained on part of the Google News dataset where the model contains 3 million words and 300-dimensional vectors ~\cite{mikolov2013distributed}. Phrases are embedded as follows. Stop words are removed using the stop word list provided by the NLTK library for the English language. The embedding of a phrase is then simply computed by taking the mean of the Google embeddings of all the words in a phrase. Given an emotion label, the closest phrase is queried by computing the cosine similarity (dot product) of the vector representing the emotion label with all of the phrase vectors. 

%% file: sections/OCC.tex
  \begin{table*}[ht]
  \caption{\label{tab:OCCPDfull}OCC-based emotional appraisals in the PD game. The ``consequences'' and ``actions of agents'' correspond to the OCC decision tree. 
  $\dSmiley$ means ``pleased'', \leftthumbsup means approving. $\heartsuit$ means desirable, and \checkmark means confirmed. Aria is ambivalent for all lines not shown. 
    For example, in the case where Aria gives while the player takes but shows regret, Aria does not disapprove of the player's action anymore (because he is showing regret), but does not actually approve of it either, so sits on the fence and does not feel admiration or reproach.}
    \begin{scriptsize}
  \begin{tabular}{|c|cc|c||l|l|l|l|l|l|l|l||ll|l|}\hline
\multicolumn{4}{|c||}{{\bf GAME PLAY}}     & \multicolumn{8}{c||}{{\bf VALENCED APPRAISALS}} & \multicolumn{3}{c|}{{\bf APPRAISED EMOTIONS}} \\ \cline{5-12}
\multicolumn{4}{|c||}{}     & \multicolumn{6}{c|}{Consequences} & \multicolumn{2}{c||}{} & \multicolumn{3}{c|}{} \\ \cline{5-10}
\multicolumn{4}{|c||}{}  & \multicolumn{2}{c|}{other} & \multicolumn{4}{c|}{self} &  \multicolumn{2}{c||}{Actions} & \multicolumn{3}{c|}{}\\ \cline{1-10}
Previous   & \multicolumn{3}{c||}{Most Recent}  & &  & \multicolumn{4}{c|}{prospects relevant?} & \multicolumn{2}{c||}{of agents} & \multicolumn{3}{c|}{} \\ \cline{1-4}\cline{7-12}\cline{13-15}
& \multicolumn{2}{c|}{Move} & Emotion & &  \multicolumn{1}{c|}{$\heartsuit$}  & \multicolumn{3}{c|}{yes} & no  & &  & \multicolumn{2}{c|}{Momentary} & Prospect-Based \\ \cline{7-10}
 &  &  &  &  & \multicolumn{1}{c|}{for}  & \multicolumn{2}{c|}{$\dSmiley$?} & & &self & other &  & &  \\
Player &  Aria & Player & Player & $\dSmiley$? &  \multicolumn{1}{c|}{other?} & future & present & \checkmark & $\dSmiley$? & \leftthumbsup? & \leftthumbsup? & Single & Compound & Single \\ \hline

give 2 & give 2  & give 2 & any & yes & yes & && & & & & happy-for & &\\ 
 & & &  & & & yes &  && & & &  & & hope\\
 & & &  & & & &yes  & yes & & & &  & & satisfaction\\
 & & &  & & &   & & & yes & yes & & joy,pride & gratification& \\
 & & &  & & &   & & & yes & & yes & joy,admiration & gratitude &\\ \hline
take 1 & give 2  & give 2 & any & yes & yes & && & & & & happy-for & &\\ 
 & & &  & & & yes & && & & &  & & hope\\
 & & &  & & & & yes  & no & & & &  & & relief\\
 & & &  & & &   && & yes & yes & & joy,pride & gratification& \\
 & & &  & & &   && & yes & & yes & joy,admiration & gratitude &\\ \hline

give 2 &  take 1 & give 2 & positive & no &no & && & & & &  pity & & \\ 
 & & &  & & & yes & && & & &  & & hope\\
 & & &  & & & & yes  & yes & & & &  & & satisfaction\\
 & & &  & & &   && & yes & & yes & joy,admiration & gratitude &\\ 
 & & &  & & &   && &  & no & & shame & & \\ \hline
take 1 &  take 1 & give 2 & positive & no &no &&  & & & & &  pity & & \\ 
 & & &  & & & yes & && & & &  & & hope\\
 & & &  & & & & yes  &yes & & & &  & & relief\\
 & & &  & & &   & & & yes & & yes & joy,admiration & gratitude &\\ 
 & & &  & & &   & & &  & no & & shame & &\\ \hline

give 2 &  take 1 & give 2 & negative & yes & no && & & & & &  gloating& & \\ 
 & & &  & & & no & && & & &  & & fear\\
 & & &  & & & & yes  & yes & & & &  & & satisfaction\\
 & & &  & & &   & & & yes & yes & & pride, joy & gratification & \\ \hline

take 1 &  take 1 & give 2 & negative & yes &no & &&  & & & &  gloating & & \\ 
 & & &  & & & no & && & & &  & & fear\\
 & & &  & & & & yes  &no & & & &  & & relief\\
 & & &  & & &   & & & yes & yes & & pride,joy & gratification & \\ \hline

give 2 &  give 2 & take 1 & no regret & no & yes & &&& & & &  resentment & &\\
& & & & & & no & & & & & & & &fear \\
& & & & & & & no & no & & & & && disappointment \\
& & & & & & & & & no & & no & distress,reproach & anger  &\\
& & & & & &  & && & yes &  & pride & &\\ \hline

take 1 &  give 2 & take 1 & no regret & no & yes & & & & & & & resentment & & \\
& & & & & & yes && & & & & & &fear \\
& & & & & & & no & yes & & & & && fears-confirmed \\ 
& & & & & & && & no & & no & distress,reproach & anger  &\\
& & & & & &  &&& & yes &  & pride & &\\\hline

give 2 &  give 2 & take 1 & regret & no & yes & && & & & & resentment && \\
& & & & & & yes & & & & & & & &hope \\
& & & & & & & no & no & & & & && disappointment  \\
& & & & & & & & & no &  & & distress & &\\
& & & & & &  &&& & yes &  & pride & &\\ \hline

take 1 &  give 2 & take 1 & regret & no & yes & & & & & & & resentment && \\
& & & & & & yes & & & & & & & &hope \\
& & & & & & & no & yes & & & & && fears-confirmed \\ 
& & & & & & &&  & no &  & & distress & &\\
& & & & & &  && & & yes &  & pride & &\\\hline

take 1 &  take 1 & take 1&any & no & no &  && & &&  &  pity && \\ 
& & & & & & no && & & & & & &fear \\
& & & & & & & no & yes & & & & && fears-confirmed \\ 
& & & & & & & & & no & no & & distress,shame & remorse &\\
& & & & & & & & & no & & no & distress,reproach & anger  &\\ \hline

give 2 &  take 1 & take 1&any & no & no &  && & & &  &  pity && \\ 
& & & & & & no & & & & & & & &fear \\
& & & & & & & no & no & & & & && disappointment  \\ 
& & & & & & & & & no & no & & distress,shame & remorse &\\
& & & & & & & & & no & & no & distress,reproach & anger  &\\ \hline

\end{tabular}
  
    \end{scriptsize}
\end{table*}

\section{OCC Model of Emotion}
\label{sec:occ}
According to the OCC model~\cite{Ortony2005}, emotions arise as a valenced reaction to the consequences of events, to the actions of agents, and to the aspects of objects. In our game situation, the aspects of objects (leading to the emotions of love and hate) do not change and so may influence overall mood but will not change substantially over the course of the interaction. We therefore focus on actions and events only. Emotions in these categories are caused by the immediate payoffs, or by payoffs looking into the future and past. Within each category, there are a number of further distinctions, such as whether focus is on the self or the other, and whether the event is positive or negative.  There are 20 emotions in the model after removing "love" and "hate".

\subsection{OCC appraisals in the Prisoner's dilemma}
Table~\ref{tab:OCCPDfull} gives the OCC interpretation of emotions in the Prisoner's Dilemma game. Each row gives the most recent move of both agent (Aria) and player (human), and the momentary emotions appraised on the {\em consequences of events} and on the {\em actions of agents}, both of which are appraised for both self and other. The prospect-based {\em consequences of events} are evaluated on each subsequent turn, leading to emotions shown in the last two columns based only on the player's previous move. Emotional intensity is not modeled, but could be added to increase realism.

Let us consider the immediate payoffs first. If Aria gets a payoff of $2$ or more, she is pleased, leading to joy when considering consequences for self where prospects are irrelevant (the gains directly lead to joy). Alternatively, for payoffs less than $2$, Aria is distressed by the loss. When considering her own actions, Aria is approving (and so feels pride) if she gives, because this seems an appropriate action in hindsight. However, if Aria takes, she disapproves of her own action because she has done something wrong, leading to shame. Unless if the player shows a negative emotion and gives 2, then Aria will approve of her action, because she predicts that it is not going well anyway.
  
  When considering the consequences for the player, if he gets $2$ or more, then Aria estimates that it is desirable for him. If he gets $2$, then Aria is pleased, and is happy-for him, otherwise she is displeased (if he gets $3$) and feels resentment. If the player gets $0$ and shows a negative emotion, Aria is pleased and gloats, but if a positive emotion is shown, Aria is displeased and feels pity. Similarly, if the player gives then Aria is approving and feels admiration, unless the player shows a negative emotion, in which case Aria is ambivalent. Aria is also ambivalent if the player takes but shows regret, otherwise she disapproves and feels reproachful.

  Prospect-based emotions arise because Aria looks into the future and predicts how things will evolve. If she gets $2$ or more, she feels pleased and hope is elicited because she estimates things will go well. However, if the player shows a negative emotion and Aria has taken, then she feels fear because she predicts a reprisal. If she gets less than $2$ she feels fear about the future, however, if the player shows regret then she feels hopeful. Looking into the past, if the player's action changed from the last time, Aria's hopes and fears are disconfirmed, leading to the prospect-based emotion of disappointment. If the player's action does not change, then Aria's hopes and fears are confirmed, leading to prospect-based emotions of satisfaction (multiple give actions) or fears-confirmed (multiple take actions).
  At the start of the game (not shown in Table~\ref{tab:OCCPDfull}), Aria feels hopeful because she is pleased about the prospect of positive payoffs in the game.

\subsection{Coping}
Once an emotion is appraised, coping is used to figure out what action to take.

Five coping strategies are taken from~\cite{Gratch2004}: acceptance, seeking support, restraint, growth, and denial, and these are applied as shown in Table~\ref{tab:coping}. At the game's start, Aria's hope leads to the support seeking coping mechanism, and thus to an initial cooperative action.

\begin{table*}[htp]
\caption{\label{tab:coping} Coping strategies for the PD bot, including last player emotion, and the last two player moves.}
 \begin{small}
  \begin{tabular}{|cc|c|l|l|c|}
    \hline
\multicolumn{2}{|c|}{Player moves} & Player & coping & & next  \\
t-2 & t-1 & emotion (t-1)& strategy & example &  move\\ \hline
take 1 & take 1 & - & acceptance:  live with bad outcome & {\em ``oh well, we're doomed''} & take 1  \\
take 1 & give 2 & positive & growth: positive reinterpretation & {\em ``this might be turning around''}& give 2 \\
take 1 & give 2 & negative & growth+denial: positive reinterpretation  & {\em ``maybe he didn't mean that emotion''}& give 2 \\
give 2 & take 1 & regret & restraint: hold back negative, keep trying & {\em ``he's a good person really''} & give 2 \\
give 2 & take 1 & not regret & denial: deny reality, continue to believe & {\em ``maybe its not so bad''} & give 2 \\
give 2 & give 2 & -  & seek support: understanding and sympathy &{\em ``Let's cooperate together on this''}& give 2 \\ \hline
\end{tabular}

 \end{small}
\end{table*}

%% file: sections/experiment.tex
\section{Experiment}

To assess how humanness of the OCC agent is perceived, we ran an experiment on Mechanical Turk, where the participants played the Prisoner's Dilemma against different agents with different strategies and emotional displays. We then asked participants to evaluate each agent on Human Nature and Human Uniqueness traits. 

\subsection{Methodology}

The experiment consisted of two parts: the Prisoner's Dilemma game and a questionnaire. Participants played 25 rounds of the game against an agent, which was randomly assigned based on the experimental condition. Afterwards, they filled out a questionnaire assessing how they perceived different aspects of humanness of the agent that they were assigned to. We ensured that the participants would pay attention and try to maximize their outcome by providing a bonus according to the points they earned in the game. The amount of the bonus was significantly larger than the initial payment. Further, participants were told that they will play up to 30 rounds because knowing the number of rounds can affect people's strategy in the final rounds.

\subsubsection{Experimental Conditions}

The experiment had three between-participant conditions. In all conditions, participants played the same number of game rounds against an opponent (Aria) and answered the same survey. However, the behavior and emotional displays of the agent changed depending on the experimental condition. The conditions of the experiment were as follows:

\begin{enumerate}
\item \textbf{OCC:} Agent acts according to Table~\ref{tab:coping}. Emotional displays are selected randomly from the set defined in Table~\ref{tab:OCCPDfull}, and facial expressions and utterances are applied as described above.
\item \textbf{Emotionless:} Agent plays tit-for-2-tats (cooperates immediately upon cooperation, but defects only after two defections), shows no emotional expressions in the face and says nothing. This agent still shows quiescent behaviours.
\item \textbf{Random:} Agent plays tit-for-2-tats. Emotional displays are randomly drawn from the set of 20 emotions, and facial expressions and utterances are selected on the basis of that.
\end{enumerate}

The Emotionless agent is added to ensure that the participants are paying attention to the emotions when rating the humanness of the agents. The Random condition enables us to study whether the participants pay attention to the differences in the emotional displays, and the relationship between their actions/emotions and the agent's emotions, when rating the humanness of the agents.

\subsubsection{Questionnaire}

We use four sets of questions before and after the game. These questions are as follows:
\begin{enumerate}
 \item \textbf{Demographic Questionnaire:} Before the game, participants were asked to provide their demographic information (i.e, age and gender). We use this information to control for possible effects of gender and age on perception of the humanness of the agents. Participants could decide not to disclose this information.

 \item \textbf{Humanness Questionnaire:} After the game, we used the humanness assessing questionnaire to assess participants' perception of agents' emotions and behaviors. The humanness questionnaire consisted of 18 questions. The first two questions asked participants to rate to what extend they thought that the agent behaved human-like/animal-like. and to what extend they thought it behaved human-like/machine-like. The following 16 questions used the traits proposed by Haslam {\em et al.}~\cite{Haslam2008} and assessed different \textit{Human Nature} and \textit{Human Uniqueness} traits in more details. 

\item \textbf{Enjoyment Question:} After the Humanness Questionnaire, participants were asked to rate how much they enjoyed playing the game. We used this question to study whether different emotional displays can affect participants' satisfaction.

\item \textbf{IDAQ Questionnaire:} After answering all other questions, participants answered the IDAQ questionnaire proposed by Waytz {\em et al.}~\cite{waytz2010sees}. The results from this questionnaire was used to account for individual differences in the general tendency to anthropomorphize.

\end{enumerate}

A continuous slider was used in all questions, except IDAQ, which uses an 11-scale, the standard scale for this questionnaire~\cite{waytz2010sees}. In addition to these questions, a total of six sanity-check questions with clear answers (e.g., ``How many 'a's are in the word ``Aria'''?'') were randomly placed among the questions in the Humanness and IDAQ questionnaires to ensure that the participants paid attention.

\subsubsection{Procedure}

Participants first signed the consent form and provided their demographic information. Then they played 25 rounds of the game against one of the agents, which was randomly assigned to them. After completing all 25 rounds of the game, participants answered the aforementioned set of questions regarding humanness natures, enjoyment, and IDAQ. Repeated participation was not allowed. We ensured that the participants saw and heard Aria, and used a browser that was compatible with our platform.

\subsubsection{Participants}

Participants were recruited on Amazon Mechanical Turk. 124 participants completed the game and the questionnaire (74 male, 48 female, 1 other, and 1 did not wish to share, age: [21,74]). The data from 6 participants (4 male and 2 female) were removed as they failed to pass the attention checks. Data from 1 participant (male) was removed as he was not able to hear the agent properly. Participants received an initial payment of \$0.7 and a bonus according to their performance in the game (\$0.05 for each point they earned). Participation was limited to residents of North America, who had completed at least 50 HITs and had a prior MTurk approval rate of 96\%. The experiment was approved by the University of Waterloo's Research Ethics Board.

\section{Results}

In this section, we will first demonstrate how playing against different agents affected perception of humanness. We then show the effects on the cooperation rate and on participants' enjoyment.

\subsection{Humanness}

\begin{figure}
 \begin{center}
    \includegraphics[width=0.75\linewidth]{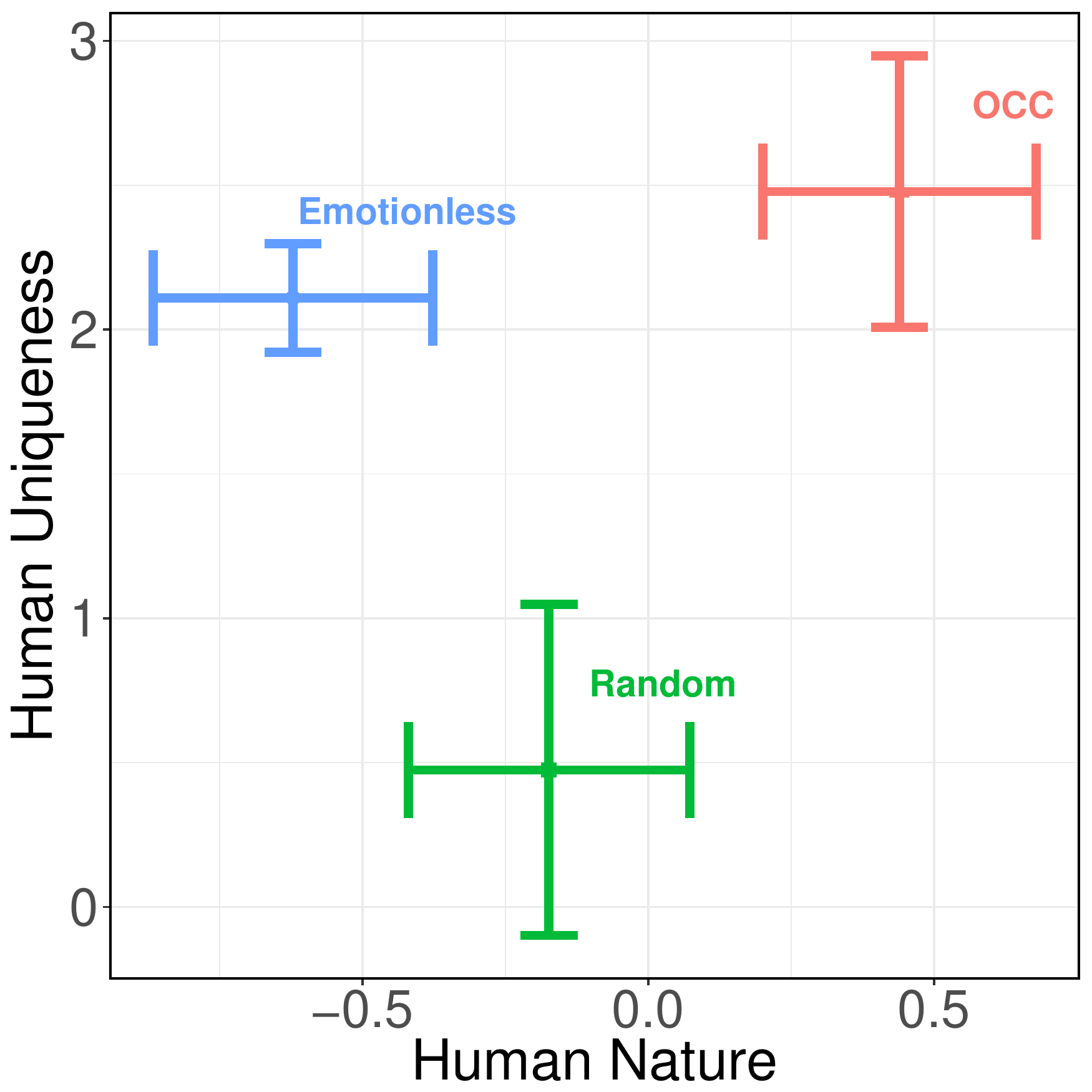}
    \end{center}
\caption{\label{fig:humanness} Rating of humanness for all conditions. X axis shows the rating of Human Nature traits (emotionality, warmth, openness, individuality, and depth) and Y axis shows the rating of Human Uniqueness traits (civility, refinement, maturity, rationality, and moral sensibility).  95\% confidence intervals are demonstrated.}
 \end{figure}

We assessed all the agents based on the rating of HN and HU traits. Figure~\ref{fig:humanness} shows the results. As hypothesized, the OCC model was perceived to be more human-like on both HN and HU aspects. We fit two linear mixed effect models predicting HN and HU ratings based on experimental condition. IDAQ, the general tendency to anthropomorphize, was controlled for. We also controled for possible effects of age, gender, and bonus (as the final bonus may affect people's perception of the agent). A random effect based on the day on which the experiment was run is fitted. The modeling results for HN and HU ratings are shown in Tables~\ref{tab:HN} and~\ref{tab:HU}, respectively. The OCC agent's HN traits were perceived to be significantly higher than the Emotionless agent, and its HU traits were perceived to be significantly higher than the Random agent. That is to say, overall, the OCC was perceived to be significantly more human-like as compared to the other two conditions. 

Further, the bonus has significantly affected perception of the HU traits, as these traits mostly describe perception of the agent's actions (e.g., rationality, sensibility). However, we did not see any effect of bonus on perception of HN traits.

\begin{table}

 \begin{small}
\begin{center} 
\caption{Linear mixed-effects model predicting the ratings for Human Nature traits based on condition. Age, gender, and anthropomorphism tendency (acquired using IDAQ) are controlled for. A random effect is fit based on the day. } 
\label{tab:HN} 
\vskip 0.12in
\begin{tabular}{lrrrr} 
\hline
Covariate &  \multicolumn{1}{c}{Estimate} & \multicolumn{1}{c}{SE} & \multicolumn{1}{c}{t} & \multicolumn{1}{c}{Pr ($>|t|$)}\\
\hline
Intercept & 0.268  &  1.931 &  0.139 & 0.890\\ 
\hdashline
Random   &  -0.434 &  0.335 &  -1.296 & 0.198 \\  
Emotionless & -0.786 &  0.343 &  -2.296 & $<0.05$ \\ 
\hdashline
IDAQ       &    0.011 &  0.008 &  1.432 & 0.155 \\   
genderFemale &   -2.870   & 1.585  &  -1.810 & 0.073 \\  
genderMale    &       -2.868  &  1.563  &  -1.835 & 0.069 \\  
genderOther &       -2.527  &  2.112  &  -1.197 & 0.234  \\  
age   &  0.046  &  0.012  &   3.813 & $<0.0005$\\
bonus    & 0.245   & 0.459 &   0.533 & 0.595 \\
\hline
\end{tabular} 
\end{center} 

 \end{small}
\end{table}

\begin{table}
\begin{small}
\begin{center} 
\caption{Linear mixed-effects model predicting the ratings for Human Uniqueness traits based on condition. Age, gender, and anthropomorphism tendency are controlled for. A random effect is fit based on the day. } 
\label{tab:HU} 
\vskip 0.12in
\begin{tabular}{lrrrr} 
\hline
Covariate &  \multicolumn{1}{c}{Estimate} & \multicolumn{1}{c}{SE} & \multicolumn{1}{c}{t} & \multicolumn{1}{c}{Pr ($>|t|$)}\\
\hline
Intercept &    10.449 & 3.708  & 2.818 & 0.006 \\
\hdashline
Random   &    -2.285 & 0.644 &  -3.551 & $<0.001$\\
Emotionless & -0.495 & 0.658 & -0.752 & 0.454 \\    
\hdashline
IDAQ    &    -0.003  &  0.015 &  -0.183 & 0.855 \\    
genderFemale     &     -2.186  &   3.045 &  -0.718 & 0.474 \\   
genderMale     &   -2.349 &   3.002 &  -0.783 & 0.436 \\  
genderOther   &    -1.269 &  4.056 & -0.313 & 0.755 \\    
age  &   0.031  &  0.023 &  1.333 & 0.185 \\
bonus     &  -2.821 & 0.882 &  -3.199 & $<0.005$\\
\hline
\end{tabular} 
\end{center} 
\end{small}
\end{table}

\subsection{Cooperation}

\begin{figure}
 \begin{center}
    \includegraphics[width=0.6\linewidth]{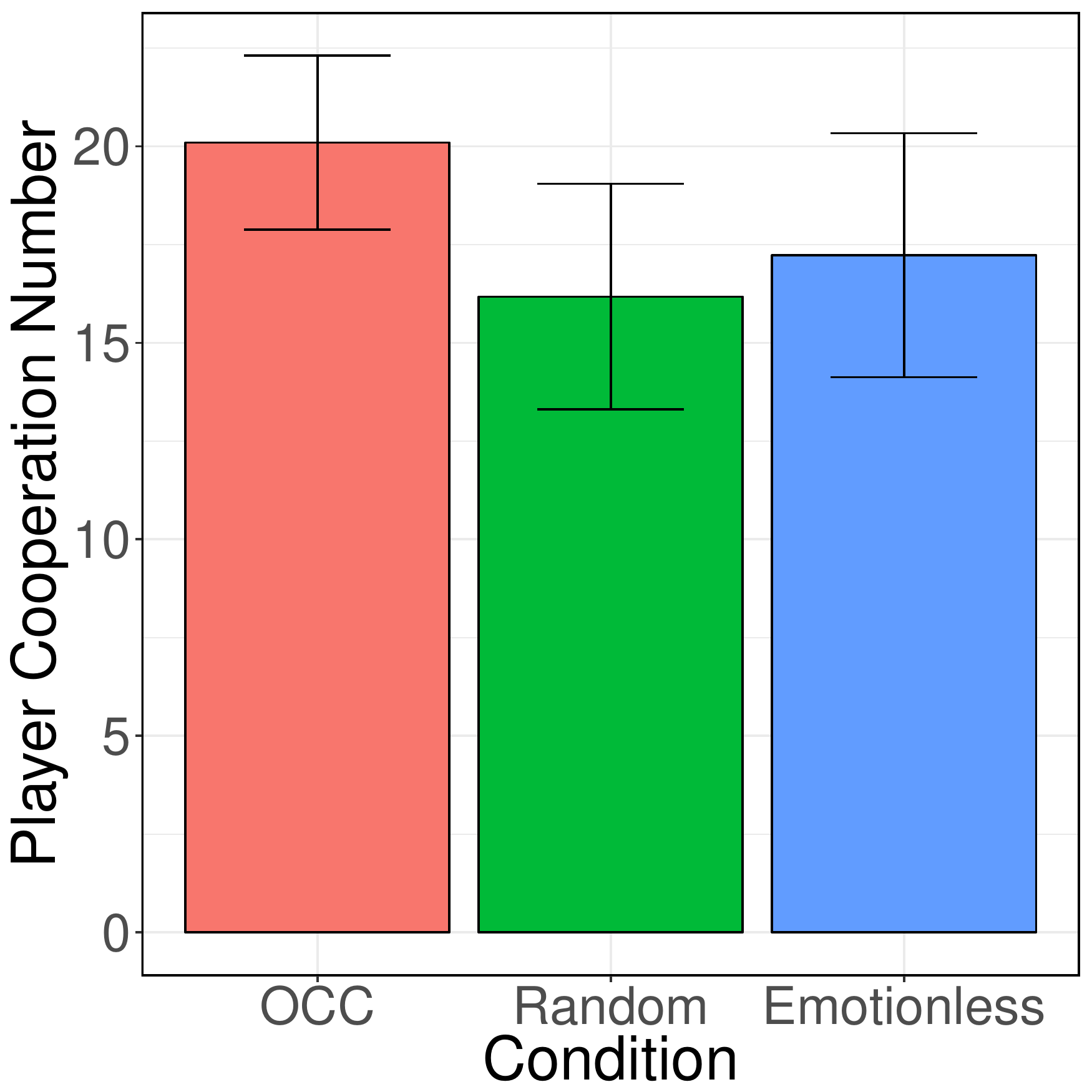}
    \end{center}
\caption{\label{fig:cooperation} Number of rounds in which the participants chose to cooperate with Aria. The maximum number of rounds were 25 for all conditions. 95\% confidence intervals are visualized.}  
\end{figure}

Next, we asked whether different emotional displays affected participants' strategies. All agents played the same strategy (i.e., tit-for-two-tats); therefore, the difference in cooperation rates among conditions can reflect the effect of the different emotional displays on participants' tendency to cooperate (in other words, trusting the agent). Figure~\ref{fig:cooperation} shows the results. OCC has the highest cooperation rate and seems to encourage cooperation. This difference is significant between the OCC and Random agent ($se= 1.851, t=-2.006, p<0.05$), however, we did not see a significant difference between the cooperation rates of the OCC and Emotionless agent.

\subsection{Enjoyment}

We know that perception of humanness of virtual agents can affect people's enjoyment in games~\cite{chowanda2016playing}. Here we ask what attributes of humans contribute to this effect. Therefore, we look into HN and HU traits independently, hypothesizing that HN traits are the key factors for enjoyment, as they distinguish humans from machines.

\begin{figure}
  \centering
  \begin{tabular}{cc}
    \includegraphics[width=0.47\linewidth]{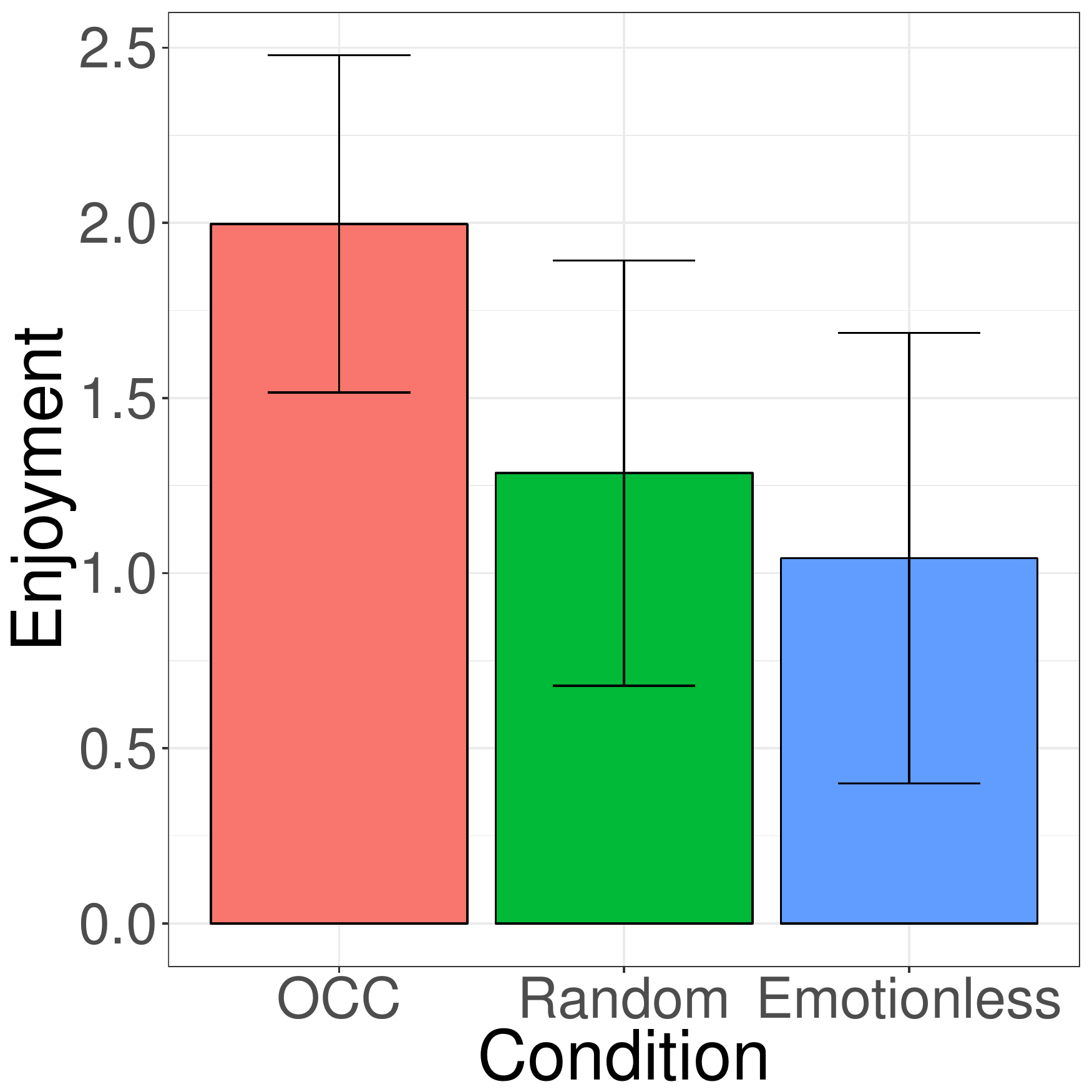} &  \includegraphics[width=0.47\linewidth]{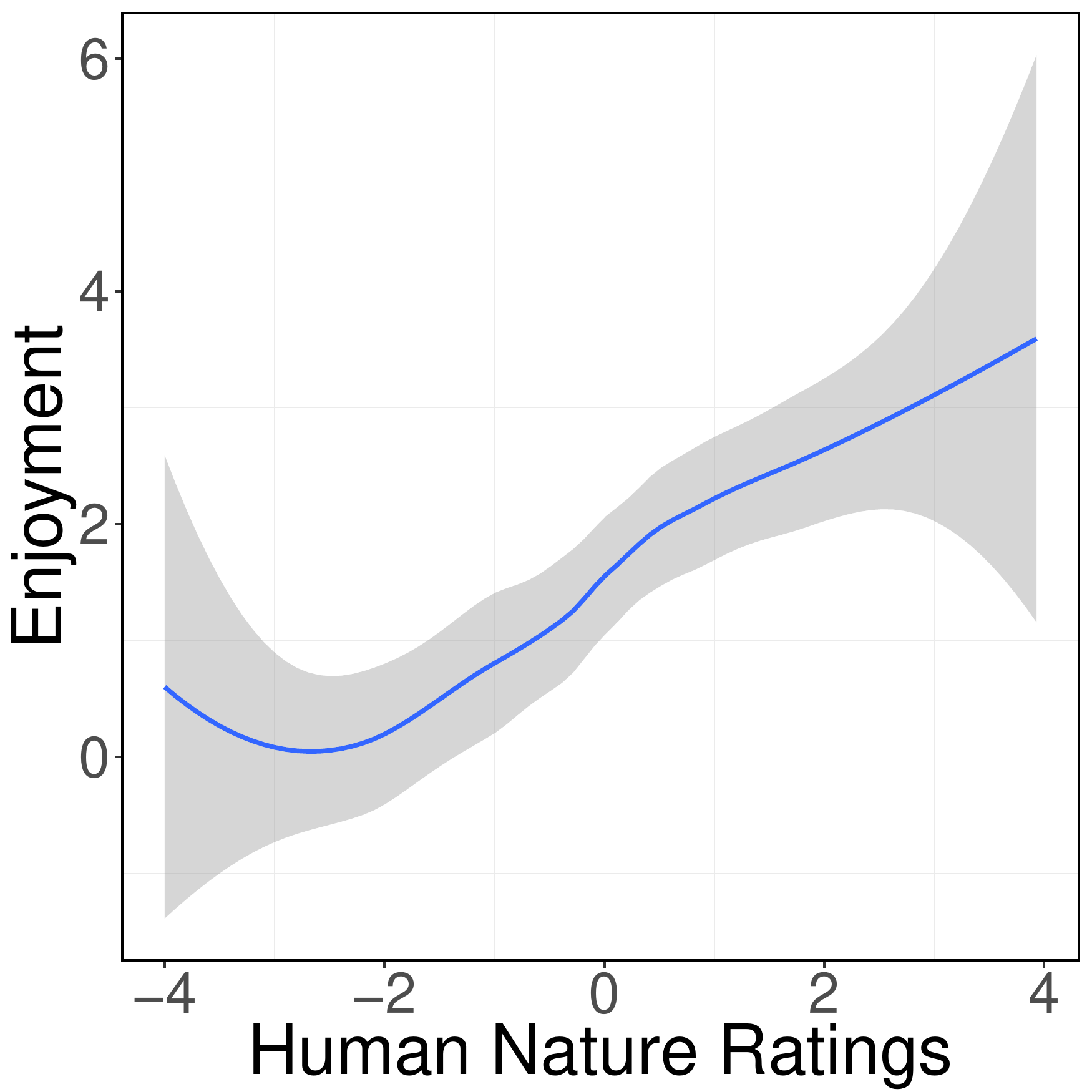} \\
    (a) & (b) 
  \end{tabular}
\caption{\label{fig:enjoymentAll} (a) Enjoyment rating for all conditions.  (b) Enjoyment rating based on perception of the HN traits. 95\% confidence intervals are visualized.}  
  \end{figure}

\begin{table}
\begin{small}
\begin{center} 
\caption{\label{tab:enjoyment} Linear mixed-effects model predicting the enjoyment ratings based on perception of HU and HN traits. Anthropomorphism tendency, and bonus are controlled for. Two random effect based on condition and day are fitted.} 
\vskip 0.12in
\begin{tabular}{lrrrr}
\hline
Covariate &  \multicolumn{1}{c}{Estimate} & \multicolumn{1}{c}{SE} & \multicolumn{1}{c}{t} & \multicolumn{1}{c}{Pr ($>|t|$)}\\
\hline
Intercept  & -1.747 &   1.308 &  -1.336 &  0.184 \\
\hdashline
HN  &  0.475 &  0.095 &   5.005 & $<0.0001$\\
HU   &  0.047 &   0.051 &   0.933 &  0.353 \\  
\hdashline
IDAQ  &  0.020 &   0.008 &  2.591 &  $<0.05$ \\  
bonus  &  0.982 &  0.484 &  2.029 &  $<0.05$ \\  
\hline
\end{tabular} 
\end{center} 
\end{small}
\end{table}

Figure~\ref{fig:enjoymentAll}(a) shows the enjoyment ratings for each condition. Playing against the OCC agent has improved users' experience of playing the game. We fit a model to look directly at how perception of HN and HU traits affects users' enjoyment in the game. Table~\ref{tab:enjoyment} shows the results. Perception of HU traits does not seem to affect users' enjoyment, however, perception of HN traits significantly affected enjoyment in the game. That is, playing against an agent that is perceived to be more human-like in HN traits (with the exact same strategy and actions) significantly increased enjoyment in the game. These results are visualized in Figure~\ref{fig:enjoymentAll}(b). As expected, the anthropomorphism tendency and the final bonus both affected happiness; therefore they are controlled for in the model.

%% file: sections/discussions.tex
\section{Discussion}

Virtual agents and assistants are used in many domains to enhance people's quality of life. We are especially interested in application of virtual agents in health care, for assisting people with cognitive disabilities such as Alzheimer's Disease in performing daily activities. Such assistants can decrease the dependence on the caregiver and reduce their burden. One challenge, however, is that the agent should be designed in a way that it can be successfully adopted by older adults with less exposure to technology. We know that affective experience increases engagement~\cite{o2008user} of users, improves loyalty~\cite{jennings2000theory}, and influences perception of humanness of the agents, which can affect people's behaviour and enjoyment~\cite{chowanda2016playing}. Therefore, in this paper, we studied how emotions affect perception of different dimensions of humanness of the agents, and users' trust.   

We utilized Haslam {\em et al.}'s definition of humanness~\cite{Haslam2008} to study how emotions affect perception of \textit{Human Nature (HN)}, distinguishing humans from machines, and \textit{Human Uniqueness (HU)}, distinguishing humans from animals. We asked how emotions affect peoples' perception of HU and HN traits of an agent, and as a result, their opinion and behaviour towards the agents. We hypothesized that although there is an emphasis on improving the HU dimension of computers (as a result of making computers more refined, rational, and moral), improving the HN dimension has seen relatively limited attention. With emotionality being an aspect of HN, we hypothesize that agents capable of showing emotions will be perceived more human-like, especially on the HN traits. 

We used a social dilemma, the prisoner's dilemma, to test this hypothesis. In prisoner's dilemma, the players cooperate if they trust the opponent, so this game enables us study how emotions and perception of humanness of agents can affect trust. Although all the agents (opponents) played the same tit-for-2-tats strategy, they differed in the emotional displays, which significantly affected perception of human-like traits. The OCC agent, capable of showing meaningful emotions, was perceived significantly more human-like on both HN and HU traits. This significantly improved participants' cooperation rate and enjoyment. \textit{Any} expression of emotion, even by the random agent, improved perception of Human Nature traits (warmth, openness, emotionality, individuality, and depth). However, displaying random emotion negatively affected Human Uniqueness traits (civility, refinement, moral sensibility, rationality and maturity), as it can make the agent look irrational and immature. That is to say, while showing proper emotions enables computer agents with Human Nature traits and fills the gap between humans and machines, showing random emotions that are not necessarily meaningful is even worse than showing no emotions for the Human Uniqueness traits, making the agent more animal-like. 

An interesting observation was that the general anthropomorphism tendency (measured through the IDAQ questionnaire) significantly and negatively affected cooperation. This may suggest an uncanny valley effect: those who anthropomorphized more perceived Aria to be more similar to humans, which resulted in disliking Aria and not trusting her~\cite{mathur2016navigating}.

 Finally, age significantly affected ratings of Human Nature traits. All the agents were perceived more human-like on the HN traits when age increased. This may be because the younger adults are more used to seeing avatars and characters in computer games, which look similar to humans, thus have a higher standard in mind regarding virtual agents. Another interesting observation was that the amount of bonus significantly affected perception of HU traits. Possibility because the participants believed that the agent (i.e. their opponent in the game) was in fact responsible for what they earned (the results), and associated a higher bonus to a better performance of the opponent (despite the fact that the strategy of all the agents was the same).
 
 Our work has a number of limitations. First, the \occ coping mechanism is theoretically difficult to justify and is usually specified in a rather ad-hoc manner~\cite{Crivelli2018}. In the simple game considered here, it provides a reasonable approximation and yields a strategy often used by humans (tit-for-two-tat). Similar coping strategies could be defined using an ``intrinsic reward'' generated by the appraisal variables. As reviewed by Broekens {\em et al.}~\cite{Moerland2017}, this intrinsic reward requires some weighting factor (e.g. $\phi$ in ~\cite{Moerland2017}) which is difficult to specify. In this simple game, we could, for example, consider that motivational relevance, which is inversely proportional to the distance from the goal, may be larger in cases where the agent predicts cooperation (e.g., the other player cooperates, or defects but shows regret), and smaller when the agent predicts defection (e.g., the other player defects and shows no regret, or defects multiple times). Motivational relevance would add intrinsic reward to the cooperation option, making it game theoretically optimal compared to defection. In the give2-take1 game, this would require adding a reward of $1$ to the cooperation option when motivational relevance was high (when there was ``hope'') and not doing so when motivational relevance was low (when there was ``fear'').
 
 A second limitation is that emotions are displayed in the face based only on a dimensional emotion model (EPA space), and neglects semantic context. For example, while {\em repentant} and {\em reverent} have different meanings which should result in different facial expressions, their EPA ratings are almost identical. Therefore there are some instances where mapping from EPA$\rightarrow$ facial expressions does seem accurate, but emotion label $\rightarrow$ facial expression not so much.
A third limitation is the limited number of emojis used to allow human players to express emotions, and the interpretation given to those emojis by the participants. While the emotion word can be seen by hovering, a better method would involve facial expression recognition to extract emotional signals directly.

%% file: sections/conclusionFuture.tex
\section{Conclusion}

This paper described our work towards understanding the effect of emotions on different dimensions of humanness of computer agents, as well as on users' cooperation tendency and enjoyment. We studied traits distinguishing humans from machines ({\em Human Nature}), and those distinguishing humans from animals ({\em Human Uniqueness}), and showed that proper expressions of emotions increases perception of human nature of agents. While researchers can successfully improve perception of {\em Human Uniqueness} traits by making agents smarter, emotions are critical for perception of {\em Human Nature} traits. This improvement also positively affected users' cooperation with the agent and their enjoyment. Further, we showed that if emotions are not reflected properly (e.g., generated randomly), they can have negative effects on perception of humanness (HU traits) and can reduce the quality of social agents, even compared to when the agent does not reflect any emotions. Therefore, it is important to find models that accurately understand and express emotions, and utilize them properly in developing virtual agents, should those agents need to be perceived as more human-like.

%% file: sections/acknowledgements.tex
\section*{Acknowledgements}

 This work was supported by Natural Sciences and Engineering Research Council of Canada (NSERC) and Social Sciences and Humanities Research Council of Canada (SSHRC) through the Trans-Atlantic Platform's Digging into Data Program. We thank Nattawut Ngampatipatpong and Sarel van Vuuren for help with the virtual human.